\documentclass[sigconf]{acmart}

\usepackage[utf8]{inputenc}
\usepackage{amsmath}
\usepackage{amsfonts}
\usepackage{float}
\usepackage{lipsum}
\usepackage{multicol}
\usepackage{xcolor}
\usepackage{natbib}
\usepackage[font=small]{caption}
\usepackage{enumitem}
\usepackage{physics,amsmath}
\usepackage{multirow}
\usepackage{booktabs}
\usepackage{natbib}
\usepackage{algorithm}
\usepackage{algorithmic}

\newcommand{\ignore}[1]{}

\AtBeginDocument{%
  \providecommand\BibTeX{{%
    \normalfont B\kern-0.5em{\scshape i\kern-0.25em b}\kern-0.8em\TeX}}}

\setcopyright{rightsretained}
\copyrightyear{}
\acmYear{}
\acmDOI{}

\acmConference[]{3rd International Workshop on Explainable AI in Finance, ICAIF 2023}{November 27,
  2023}{NY}
\acmBooktitle{3rd International Workshop on Explainable AI in Finance, ICAIF 2023} 
\acmPrice{}
\acmISBN{}

\begin{document}

\title{Privacy-Preserving Algorithmic Recourse}

\author{Sikha Pentyala}
\email{sikha@uw.edu}
\affiliation{%
  \institution{University of Washington Tacoma}
  \city{Tacoma}
  \country{USA}}

\author{Shubham Sharma}
\email{shubham.x2.sharma@jpmchase.com}
\affiliation{%
  \institution{J.P. Morgan AI Research}
  \city{New York}
  \country{USA}}
  
\author{Sanjay Kariyappa}
\email{sanjay.kariyappa@jpmchase.com}
\affiliation{%
  \institution{J.P. Morgan AI Research}
  \city{Palo Alto}
  \country{USA}}

\author{Freddy Lecue}
\email{freddy.lecue@jpmchase.com}
\affiliation{%
  \institution{J.P. Morgan AI Research}
  \city{New York}
  \country{USA}}

\author{Daniele Magazzeni}
\email{daniele.magazzeni@jpmorgan.com}
\affiliation{%
  \institution{J.P. Morgan AI Research}
  \city{New York}
  \country{USA}}

\renewcommand{\shortauthors}{Pentyala, et al.}

%
%
\renewcommand{\algorithmicrequire}{\textbf{Input:}}
\renewcommand{\algorithmicensure}{\textbf{Output:}}
\newcommand{\purple}[1]{\textcolor{purple}{Questions: #1}}
\newcommand{\TODO}[1]{\textcolor{red}{TODO: #1}}
\newcommand{\shubham}[1]{\textcolor{magenta}{Shubham: #1}}
\newcommand{\sikha}[1]{\textcolor{blue}{Sikha: #1}}
\newcommand{\notes}[1]{\textcolor{brown}{#1}}
\newcommand{\CommentLine}[1]{
    \State // \textit{#1}
}
\newcommand{\xmark}{$\textcolor{red}{\times}$}%

\begin{abstract}
  When individuals are subject to adverse outcomes from machine learning models, providing a recourse path to help achieve a positive outcome is desirable. Recent work has shown that counterfactual explanations - which can be used as a means of single-step recourse - are vulnerable to privacy issues, putting an individuals' privacy at risk. Providing a sequential multi-step path for recourse can amplify this risk. Furthermore, simply adding noise to recourse paths found from existing methods can impact the realism and actionability of the path for an end-user. In this work, we address privacy issues when generating realistic recourse paths based on instance-based counterfactual explanations, and provide PrivRecourse: an end-to-end privacy preserving pipeline that can provide realistic recourse paths. PrivRecourse uses differentially private (DP) clustering to represent non-overlapping subsets of the private dataset. These DP cluster centers are then used to generate recourse paths by forming a graph with cluster centers as the nodes, so that we can generate realistic - feasible and actionable - recourse paths.  We empirically evaluate our approach on finance datasets and compare it to simply adding noise to data instances, and to using DP synthetic data, to generate the graph. We observe that PrivRecourse can provide paths that are private and realistic.
\end{abstract}



\maketitle

%
%
\section{Introduction}

Numerous financial systems, such as credit approval processes, are driven  by Machine Learning (ML) models to provide decisions. When users are adversely affected by these decisions, it becomes crucial to offer transparent explanations. This also aligns with regulatory requirements like ECOA, FCRA \cite{ammermann2013adverse},  the 'Right to Explanation' in the EU-GDPR \cite{goodman2017european} and the US-AI Bill of Rights \cite{aibillofrights}.
These explanations can be in the form of sequential steps to get preferred or favorable outcomes. Such recommended steps are considered as providing algorithmic recourse to the affected user. Single-step recourses are often computed through counterfactual explanations (CFEs) \cite{karimi2022survey}, which are a type of explanation that suggest the changes that can be made to the input to get a different decision \cite{wachter2017counterfactual}. Recent work has shown that providing a single-step path may not be enough: a given recourse should constitute multi-step changes toward a favorable outcome \cite{verma2022amortized, venkatasubramanian2020philosophical}. Additionally, it is important that such recourse paths should be realistic in nature (i.e. feasible and actionable) for it to be useful for the end-users.
\begin{figure}[h]
    \centering
    \includegraphics[width=0.5\textwidth]{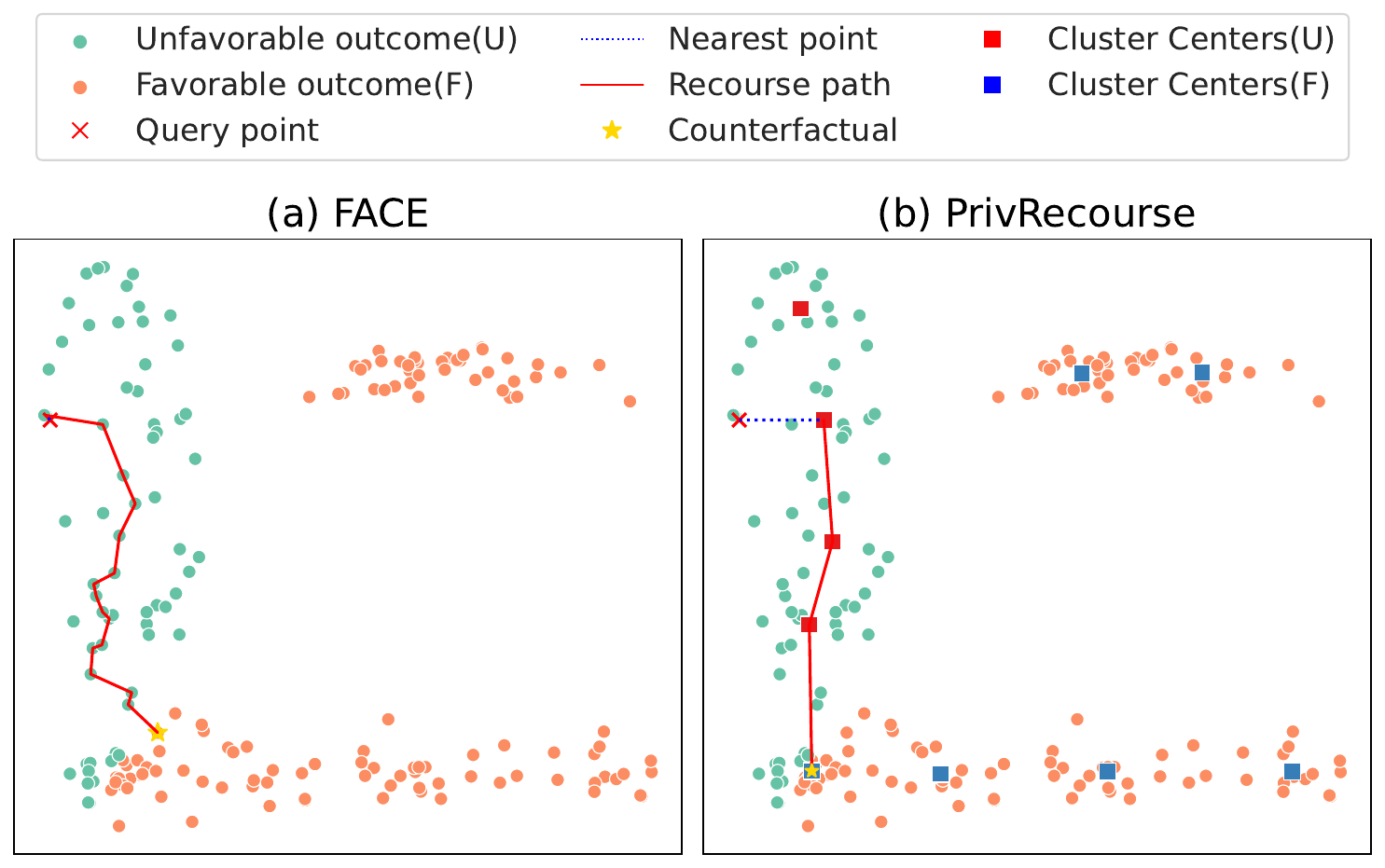}
    \caption{ (a) The recourse path generated by non-private FACE passes through the denser data manifold giving a realistic recourse path and counterfactual explanation (CFE). (b) We propose PrivRecourse, which generates privacy-preserving cluster centers that can be published with ($\epsilon,\delta$)-DP guarantees which can then be used to generate privacy-preserving recourse path and CFE.}
    \label{fig:intro}
\end{figure}
But providing a CFE - a single-step recourse path - is challenging when private training data is involved as
CFEs can inadvertently expose sensitive information of the users in the training dataset to adversaries \cite{sokol2019counterfactual} violating privacy regulations such as RFPA, GDPR, CCPA, and AI Bill of Rights which aim to safeguard individuals' data privacy.  Adversaries can exploit membership attacks, such as counterfactual-distance-based attacks \cite{pawelczyk2023privacy} or explanation linkage attacks \cite{goethals2022privacy}, to identify individuals within the training dataset and put their privacy at risk. These risks are heightened when using instance-based explanations that return examples from the training data itself \cite{wexler2019if, poyiadzi2020face,brughmans2023nice} to form a recourse and the final counterfactual with a favorable outcome. Providing several instances from the data for a multi-step recourse path further heightens this concern.

\begin{table*}[h]
    \centering
    \small{
    \begin{tabular}{l || c | c | c | c | c }
        Work & CFE Method  &  Private BB Model &  Private CFE & Realistic recourse path & Private recourse path \\
        \hline
        \cite{mochaourab2021robust} & \cite{wachter2017counterfactual} & \xmark &  \textcolor{teal}{\checkmark} &  \xmark & \xmark \\
        \cite{yang2022differentially} & Sim. to \cite{van2021interpretable} & \xmark &  \textcolor{teal}{\checkmark} &  \xmark & \xmark \\
        \cite{goethals2022privacy} & \cite{mothilal2020dice} & \xmark &  \xmark\footnotemark &  \xmark  & \xmark \\
        \midrule
        \cite{hamer2023simple} & \cite{hamer2023simple} & \xmark &  \xmark &  \xmark & \xmark \footnotemark \\
        \cite{huang2023accurate} & \cite{wachter2017counterfactual} & \xmark &  \textcolor{teal}{\checkmark} &  \xmark & \xmark\\
        {\textbf{\texttt{PrivRecourse} [Our work]}} & \cite{poyiadzi2020face} & \textcolor{teal}{\checkmark} &  \textcolor{teal}{\checkmark} &  \textcolor{teal}{\checkmark} & \textcolor{teal}{\checkmark} \\
        
    \end{tabular}
    }
    \caption{Works on privacy-preserving CFE and recourse. Private BB Model implies a model agnostic method. \emph{Private} refers to differentially private algorithms. \textsuperscript{1}Privacy notion of k-anonymity.\textsuperscript{2}Propose private recourse directions and not paths.}
    \label{tab:rw}
\end{table*}
\vspace{0.2em}
\noindent
\textbf{Our goal.} The goal of our work is to develop an end-to-end differentially private~\cite{dwork2014algorithmic} pipeline for generating realistic counterfactuals and recourse paths. While recent works have proposed methods to provide privacy-preserving CFEs with single-step recourse, to the best of our knowledge, there is no work that provides realistic multi-step recourse paths with privacy guarantees. We aim to address this gap in literature through this work. 


\vspace{0.2em}
\noindent
\textbf{Baseline methods.}
We develop simple baseline methods by adapting the FACE~\cite{poyiadzi2020face} method to make it differentially private. The FACE  method provides realistic CFEs and recourse by building a graph on training instances. The recourse path provided by FACE consists of samples from the training dataset, thus directly revealing the individuals' data. We consider two simple methods to make FACE differentially private, which act as the baselines for our proposal:
\begin{itemize}
    \item \emph{Differentially private (DP) training data:} Instead of running FACE directly on the original data, we can release the differentially private training instances or prototypes as obtained using \cite{kim2016examples} - by adding noise to individual training samples. A graph constructed using these noisy training samples can then be used with FACE to find recourse paths.
    \item \emph{Synthetic training data:} Another approach consists of generating synthetic data with differential privacy and then applying the FACE method on the graph formed with this synthetic data.
\end{itemize}
Through our empirical studies, we find that the noisy data produced by both of these baseline methods do not sufficiently preserve the distribution of the training data. Consequently, they can produce low-quality recourse paths that may be unrealistic.

\vspace{0.2em}
\noindent
\textbf{Our approach.}
We propose \emph{PrivRecourse} to generate more realistic recourse paths with DP guarantees. Our key insight is that in the presence of noise, clustering better preserves the structure of the data, enabling us to produce realistic recourse paths that can have nodes along the dense regions of the data (Fig. \ref{fig:intro}(b)). We leverage this insight by first partitioning the training data into non-overlapping subsets with differentially private clustering. The cluster centers generated in this process represent a private version of the training data. We can use the graph constructed on these cluster centers with FACE to produce privacy-preserving recourse paths that are more realistic.
To have an end-to-end privacy-preserving pipeline, we consider differentially private black-box machine learning models and make any of the pre-processing techniques for data differentially private. In this way, we obtain differentially private information that can be published to get recourse paths for any number of query instances. Then, we use the PrivRecourse method to generate realistic recourse paths. The key contributions of our paper are as follows:
\begin{itemize}[leftmargin=*,noitemsep,topsep=0pt]
\item We propose a privacy-preserving pipeline for \textit{generating realistic recourse paths} for an \textit{unlimited number of query instances} while providing privacy guarantees to the individuals in the training dataset. 
\item We consider differentially private clustering and show that forming a graph on differntially private cluster centers can provide \textit{realistic recourse paths}.
\item We empirically analyze the \textit{impact of privacy on the realism of the recourse path} for the proposed approach and compare it to the baseline solutions that use synthetic training data and DP training data. We perform this study on openly available finance datasets. 
\end{itemize}
\section{Related Work}

The field exploring the intersection between privacy and explainability is nascent. Attack models have been designed to show how explanations can be leveraged to infer membership in the training data or magnify model inversion or reconstruction attacks\cite{chen2023dark, pawelczyk2023privacy, goethals2022privacy, duddu2022inferring,naretto2022evaluating, shokri2021privacy,zhao2021exploiting,slack2021counterfactual,milli2019model}. Most of the work that provide privacy-preserving explanations focus on feature-attribution based methods\cite{patel2022model, datta2016algorithmic, harder2020interpretable, li2023balancing, wang2019interpret, nguyen2022xrand}. Limited works are available in the literature addressing privacy-preserving CFEs and recourse. We summarize them in Tab. \ref{tab:rw} to showcase the need for PrivRecourse. 

\noindent
\subsection{Privacy-preserving counterfactual explanations.}
Mochaourab et al. \cite{mochaourab2021robust} train a differentially private SVM model and propose techniques to generate counterfactual explanations that are robust. 
Montenegro et al. \cite{montenegro2021privacy} consider privacy for case-based explanations for images with a focus on realistic counterfactual explanations. They propose a GAN based approach without any differential privacy guarantees and thus are prone to membership based attacks.
Yang et al. \cite{yang2022differentially}  train a differentially private auto-encoder to generate DP prototypes for each class label. They then optimize the amount of perturbation to be made to the instance query that minimizes the distance between the query instance and the CFE,  regularizes the perturbation towards the favourable class prototype and encourages a favourable prediction.
To the best of our knowledge, we are the first to propose a fully privacy-preserving pipeline with DP guarantees to provide a CFE (the final step in the recourse path) 
using the black-box ML model and studying the impact of privacy on the quality of the realistic CFE. 

\noindent
\subsection{Privacy-preserving algorithmic recourse.}
Few works consider providing privacy-preserving recourse explicitly (and none consider the objective of having a path adhering to the data manifold). \texttt{StEP} \cite{hamer2023simple} 
provide data-driven recourse directions and mention that these recourse directions can be made private, but do not provide any private recourse paths. Recent work \cite{huang2023accurate} considers generating privacy-preserving recourse with differentially private logistic regression model without providing the multi-step path for recourse. 



%
%

\vspace{2em}
\section{Preliminaries }

\noindent
\subsection{Algorithmic recourse.}
Counterfactual explanations have been popularly considered as a means of one-step recourse and are one of the ways to explain the predictions of a model. For an input instance $\vectorbold{Z}$ and model $f$, the original formulation\cite{wachter2017counterfactual} considers counterfactuals to be the smallest changes $\Delta$ required in the input i.e.~$\vectorbold{Z}^* = \vectorbold{Z} + \Delta$ to change the prediction of the model such that $f(\vectorbold{Z}) \neq f(\vectorbold{Z}^*)$. The recourse path {\tt P} here consists of a single-step change from $\vectorbold{Z}$ to $\vectorbold{Z}^*$.  This simple single-step scheme can produce counterfactuals/recourse paths that are unrealistic, degrading the utility of recourse \cite{venkatasubramanian2020philosophical}. 

In order for the counterfactual explanation to be useful as a recourse path, the generated counterfactual should involve changes that are feasible and actionable \cite{guidotti2022counterfactual}. 
A variety of algorithms have been proposed in the literature to generate counterfactuals that satisfy actionability and other additional properties \cite{guidotti2022counterfactual,karimi2022survey,verma2020counterfactual}. Some of these methods produce a counterfactual along with a multi-step recourse path {\tt P} consisting of a sequence of points $\{\vectorbold{Z}_1, \vectorbold{Z}_2, \ldots, \vectorbold{Z}_p\}$ with $\vectorbold{Z}_p = \vectorbold{Z}^*$, such that each point in {\tt P} suggests incremental changes in the attributes towards the favorable outcome. One popular method that does so is FACE. Poyiadzi et al.\shortcite{poyiadzi2020face} note that, for the recourse path {\tt P} to be useful, each point $\vectorbold{Z}_i\in \texttt{P}$ should be feasible. To this end, they propose the FACE algorithm to produce a realistic recourse path.
\noindent
\subsection{Differential privacy (DP).} The notion of DP guarantees plausible deniability regarding an instance being present in a dataset, hence providing privacy guarantees. Consider two neighboring datasets $D$ and $D'$ that differ in a single instance, i.e.~$D'$ can be obtained either by adding or removing an instance from $D$ or vice-versa\footnote{We assume that each entry in the dataset is independent of other instances.}. A differentially private randomized algorithm $\mathcal{A}$ ensures that it generates a similar output probability distribution on $D$ and $D'$~\cite{dwork2014algorithmic}.
Formally, a randomized algorithm $\mathcal{A}$ is called $(\epsilon, \delta)$-DP if for all pairs of neighboring sets $D$ and $D'$, and for all subsets $O$ of $\mathcal{A}$'s range,
\vspace{-3pt}
\begin{equation}\label{DEF:DP}
\mbox{P}(\mathcal{A}(D) \in O) \leq e^{\epsilon} \cdot \mbox{P}(\mathcal{A}(D') \in O) + \delta.
\end{equation}
where $\epsilon$ is the privacy budget or privacy loss and $\delta$ is the probability of violation of privacy. 
The smaller these values, the stronger the privacy guarantees. 
The above description of the DP notion is under the \textit{global} model which assumes that a trusted curator exists who has access to $D$. 
\noindent
\subsection{Problem Statement.}
Consider a private dataset $D$ with $N$ training samples and $D_i = \langle \vectorbold{X} = (\vectorbold{X}^C_{i},\vectorbold{X}^O_{i}),y_{i} \rangle$ representing the $i^{th}$ training sample in $D$ where $\vectorbold{X}^C_{i}$ is the set of $n_c$ continuous feature values, $\vectorbold{X}^O_{i}$ is the set of $n_o$ categorical feature values and $y_{i}$ is the value of a class label in $\mathcal{Y}$. The data holder publishes a black-box machine learning model $f: \mathbb{R}^{n_c+n_o} \rightarrow \mathcal{Y}$ trained on $D$,  which outputs only the predicted label. For a query instance $\vectorbold{Z}$ that receives an unfavorable outcome from the model, the data holder uses an explanation algorithm $\phi(f, D)$ to provide counterfactual  $\vectorbold{Z^*}$ that produces a favorable outcome and a recourse path {\tt P} from $\vectorbold{Z}$ to $\vectorbold{Z^*}$.
{\tt P} contains a sequence of points $\vectorbold{Z}_1, \vectorbold{Z}_2, \ldots, \vectorbold{Z}_p$ with $\vectorbold{Z}_p = \vectorbold{Z}^*$, such that each point in {\tt P} suggests incremental changes in the attributes towards the favorable outcome. 
Our goal is to develop a method that generates realistic counterfactuals ($\vectorbold{Z}^*$) and recourse paths ({\tt P}) while providing $(\epsilon, \delta)$-DP for the training dataset $D$.
\section{Method}\label{sec:method}
We consider two phases in PrivRecourse. The setup and training phase trains an ML model, constructs a graph, and identifies the set of candidate counterfactuals with DP guarantees. The inference phase provides privacy-preserving multi-step recourse for a given query instance.
\noindent
\subsection{Training phase}
Our proposed pipeline outlined in Algorithm \ref{alg:main} comprises of the following main steps:
\begin{enumerate}[leftmargin=*,noitemsep,topsep=0pt]
    \item Train a DP ML model $f$ on $D$ with a privacy budget of $(\epsilon_{f}, \delta_{f})$. Publish $f$ as a black-box model that outputs the value of \texttt{predict()} only. Any number of queries can be made on $f$ while preserving the privacy of $D$ under $(\epsilon_{f}, \delta_{f})$-DP guarantees (Line 1 in Algorithm \ref{alg:main}). 
    \item Run an unsupervised clustering algorithm to partition $D$ into $K$ non-overlapping subsets $D^{(1)},D^{(2)},\ldots,D^{(k)},\ldots,D^{(K)}$ with $(\epsilon_{k}, \delta_{k})$-DP guarantees (Line 2 in Algorithm \ref{alg:main}). Each subset, $D^{(k)}$, is represented by its cluster center $C_k$, which is equivalent to the DP representation of the subset.
    \item Construct a graph $G = (\mathcal{V},\mathcal{E})$ where $\mathcal{V}$ is the set of cluster centers  $\mathcal{C} = \{C_k | k \in \{1 \ldots K\}\}$ (Line 3 -- 18 in Algorithm \ref{alg:main}). The nodes $\mathcal{V}$ are connected by an edge if they do not violate constraints thus providing actionability. The value of weights of the edges is computed based on the distance between the nodes and their density scoring given by the density estimate of $\mathcal{C}$. $G$ is published with $(\epsilon_{k}, \delta_{k})$-DP due to post-processing property.  
    This provides closeness to data manifold and feasibility. To get closer to the training datafold $D$, one could also opt for computing the weights based on the density estimate of $D$ by publishing the privacy-preserving density estimate on $D$ \cite{wagner2023fast}.  
    \item Given the favourable outcome $y$, get a set of candidate counterfactuals $\mathcal{Z}_{CF} \subset \mathcal{V}$ such that for each $\vectorbold{Z_k} \in \mathcal{Z}_{CF}$, $f(\vectorbold{Z_k}) = y$ (Line 13 -- 17 in Algorithm \ref{alg:main}). Publish $\mathcal{Z}_{CF}$.
\end{enumerate}

The above process is equivalent to publishing the explanation algorithm $\phi(f, G)$ under $(\epsilon_{f}+\epsilon_{k}, \delta_{f}+\delta_{k})$-DP guarantees due to sequential composition.

\begin{algorithm}[tb]
\caption{PrivRecourse - Algorithm to publish privacy-preserving graph and candidate counterfactuals}
\label{alg:main}
{
\begin{algorithmic}[1] 
\REQUIRE Dataset $D$, Number of clusters to form $K$, privacy budget $\epsilon$, favourable outcome $y$, Number of iterations $T$, distance function $d()$, distance threshold $d_{th}$\\
\ENSURE $f(), G$ and $\mathcal{Z}_{CF}$
\STATE Train a ML model $f$ with $(\epsilon_{f}, \delta_{f})$-DP privacy guarantees, publish $f$.
\STATE Compute $\mathcal{C} \gets ConvergentDPCluster(K,D,T,\epsilon_{k})$ with $(\epsilon_{k}, \delta_{k})$-DP privacy guarantees (ref. \cite{lu2020differentially}).
\STATE Compute density estimate $\rho^\mathcal{C}$.
\STATE Consider weight function $w_{ij} \propto \frac{d(C_i,C_j)}{\rho((C_i + C_j)/2)}$
\FORALL{$C_i,C_j$ in $\mathcal{C}$} 
\IF{$d(C_i,C_j) < d_{th}$ and all constraints satisfied}
\STATE Connect a bidirectional edge between $(D_i,D_j)$ with weight $w_{ij}$
\ELSIF{$d(C_i,C_j) < d_{th}$ and constraints satisfied in one direction}
\STATE Connect a unidirectional edge between $(D_i,D_j)$ with weight $w_{ij}$ \\ \textit{Note that if no constraints are satisfied, either construct no edge or construct an edge with $w_{ij} = \infty$.}
\ENDIF
\STATE $G \gets G \cup w_{ij}$
\ENDFOR
\FORALL{$C_i$ in $\mathcal{C}$} 
\IF{$f(C_i) = y$}
\STATE $\mathcal{Z}_{CF} \gets \mathcal{Z}_{CF} \cup C_i$
\ENDIF
\ENDFOR
\STATE Publish $G$ and $\mathcal{Z}_{CF}$
\end{algorithmic}
}
\end{algorithm}


\begin{algorithm}[tb]
\caption{PrivRecourse  - Algorithm to get privacy-preserving recourse}
\label{alg:inference}
{
\begin{algorithmic}[1] 
\REQUIRE: Query instance $\vectorbold{Z}$, published information $f(), G$ and $\mathcal{Z}_{CF}$ favourable outcome $y$,distance function $d()$ \\
\ENSURE: \texttt{P} and $\vectorbold{Z^*}$ 
\STATE $\mathcal{Z}_{F} \gets V - \mathcal{Z}_{CF}$
\STATE $\vectorbold{Z_1} \gets \underset{\vectorbold{Z_f} \in \mathcal{Z}_{F}}{\mathrm{argmin}}(d(\vectorbold{Z},\vectorbold{Z_f}))$
\FORALL{$\vectorbold{Z_{cf}} \in \mathcal{Z}_{CF}$}
\STATE Run Djikstra's algorithm to get the shortest path from $\vectorbold{Z_1}$ to  $\vectorbold{Z_{cf}}$.
\ENDFOR
\STATE Consider the shortest path among all computed paths as \texttt{P} and the last node in the path as $\vectorbold{Z^*}$.
\STATE \textbf{return} \texttt{P} and $\vectorbold{Z^*}$ 
\end{algorithmic}
}
\end{algorithm}

\noindent
\textit{Description of Step 1.}
We assume that a DP ML model has been trained and tuned to obtain desired utility. We propose to use DP-SGD\cite{abadi2016deep}, PATE\cite{papernot2018scalable} or other privacy-preserving methods\cite{chaudhuri2011differentially,grislain2021dp} that provide DP guarantees and offer utility for the selected ML model. To make the ML model end-to-end differentially private, we propose to make any preprocessing steps on the dataset also differentially private. We can either assume domain knowledge for feature preprocessing as we do in our empirical analysis or perform privacy-preserving feature preprocessing using Randomized Response or Lapalce or Guassian Mechanisms \cite{dwork2014algorithmic}. 

\noindent
\textit{Description of Step 2.} 
We partition $D$ into clusters (non-overlapping subsets) such that their cluster centres  represent the data manifold of $D$. We employ a privacy-preserving clustering algorithm inspired by \cite{lu2020differentially, joshi2023k}, which ensures convergence within a specified privacy budget per iteration. 
The key idea is adding bounded noise to the cluster centers using the exponential mechanism during Lloyd's KMeans clustering. First, random points are chosen as initial centers  $\mathcal{C}_0$. Each $D_i$ is then assigned to the nearest center  $C_k$ on \textit{l2} distance. Then, for each cluster $D_k$ at the $t^{th}$ iteration, a convergent zone is created around the current $C^t_k$ and previous centers $C^{t-1}_k$. Subzones are formed in these zones using KMeans. A subzone is sampled based on number of data points the subzones contain. Next, a data point is sampled from the sampled subzone using the exponential mechanism as the new cluster center $C^{t+1}_k$. The exponential mechanism uses distance as the scoring function with a sensitivity of 1. After the algorithm converges, Laplace noise is added to  to the count of training data points in each cluster and then mean is computed using the DP counts.

\noindent
\subsection{Inference phase.}
To get a counterfactual explanation for a given query instance $\vectorbold{Z}$ with PrivRecourse, we first get the nearest node $\vectorbold{Z_1}$ to $\vectorbold{Z}$ defined by the given distance function such that $\vectorbold{Z_1} \in G, f(\vectorbold{Z}_1) \neq y$ (Line 1 -- 2 in  Algorithm \ref{alg:inference}).
We then get the shortest distances between $\vectorbold{Z}$ and candidates in $\mathcal{Z}_{CF}$ 
The candidate $\vectorbold{Z^*}$ with the minimum shortest distance is the privacy-preserving counterfactual (Line 3 -- 7 in  Algorithm \ref{alg:inference}) and the shortest path from $\vectorbold{Z_1}$ to $\vectorbold{Z^*}$ is the privacy-preserving recourse path {\tt $P$} 
suggested along the data manifold. All points $\vectorbold{Z}_i$ in {\tt $P$} except $\vectorbold{Z^*}$ satisfy $f(\vectorbold{Z}_i) \neq y$.  As the $\phi(f, G)$ is published, any number of counterfactuals explanations and recourse paths can be queried where the answers to the queries provide $(\epsilon_{f}+\epsilon_{k}, \delta_{f}+\delta_{k})$-DP guarantees due to the post-processing property of DP.


\begin{table*}[h!]
    \caption{Evaluation of the privacy-preserving recourse and CFE. The values are averaged over \#CFE (number of CFE's) for which the recourse path is generated and also averaged over 3 runs. 
    FACE is the non-private method. The baseline approach adds random noise to the individual samples, DP-SD uses DP synthetic data, Ours is the proposed approach PrivRecourse. $\epsilon_k=1$ and maximum iterations of $10$ are used to keep the privacy budget the same across all methods. Feas.=Feasibility, Prox.=Proximity, Spar.=Sparsity, Rob.=Robustness, T(s)=Average time in seconds, Den=\texttt{PDensity},  Dist=\texttt{PDistanceManifold},  Red.=Redundancy} 
    \centering
    \scalebox{1}{
    {
        \begin{tabular}{ cll  | cccc | cccc }
        \toprule
        &&  &    \multicolumn{4}{c}{ADULT} & \multicolumn{4}{c}{HELOC}  \\
        &&  &    \multicolumn{4}{c}{(\#CFE: 266 , k=500)} & \multicolumn{4}{c}{(\#CFE: 296, k=500)} \\
        \midrule
        &&  &  {FACE} & {Baseline} & {DP-SD} & {Ours} & {FACE} & {Baseline} & {DP-SD} & {Ours} \\
        \midrule

        \parbox[t]{0.5mm}{\multirow{5}{*}{\rotatebox[origin=c]{90}{\textbf{Rec. Path}}}}&\multirow{3}{*}{Feas.}  
                & Den $\uparrow$  & 2.38  & -0.34 & 2.15 & \textbf{2.25} 
               & 2.56  & -0.05 & 2.53 & \textbf{2.54}\\
        &        & Dist $\downarrow$ & 0  & 0.21 & 0.15 & \textbf{0.01}  
                              & 0 & 0.38 & 0.05 & \textbf{0.03}\\
        
        \cmidrule{2-11}

        &\multirow{2}{*}{Prox.}
        
        & PL1 $\downarrow$&  0.07 & 0.78 & \textbf{0.09} & 0.17  
                      & 0.12  & 0.67 & \textbf{0.13} & 0.18\\
        
        && PL2 $\downarrow$& 0.05  & 0.43  & \textbf{0.07} &  0.11 
                      & 0.06  & 0.34 & \textbf{0.07} & 0.10\\     

        \cmidrule{2-11}
                &\multirow{1}{*}{Spar.}
        & PL0 $\downarrow$&  1.72 & 5 & \textbf{2.36} &  3.79 
                      & 4.51  & 5 & \textbf{4.36} & 4.98\\

        \midrule
        \parbox[t]{0.5mm}{\multirow{3}{*}{\rotatebox[origin=c]{90}{\textbf{CFE}}}}&\multirow{1}{*}{Spar.}
        

        & Red. $\downarrow$& 0.43  & 3.23  & \textbf{1.27} &  1.45 
                      & 1.56  & 2.56 & \textbf{2.21} & 2.29\\
         \cmidrule{2-11}
        &\multirow{1}{*}{Rob.}
        
        & yNN $\uparrow$& 0.79  & 0.13 & 0.78 &  \textbf{0.94 }
                      & 0.67  & 0.40 & 0.67 & \textbf{0.74}\\
        \midrule

        &\multicolumn{1}{c}{T(s)$\downarrow$} &
        
                      &  75.79 & 70.98  & 72.30 &   \textbf{31.96}
                      & 11.70  & 6.45 & 10.30 & \textbf{4.40}\\
        
        \bottomrule
        \end{tabular} 
        }
        }
    \label{tab:results}
\end{table*}




%
%
\section{Experiments}

\subsection{Datasets and model training.}
We evaluate PrivRecourse on Adult \cite{misc_adult_src} and HELOC \cite{misc_heloc_src} datasets. The Adult dataset has 32561 training samples and 16281 test samples. The task is to determine whether a person makes over 50K a year. We consider 'age', 'education-num', capital-gain','capital-loss','hours-per-week' as predictors for this task, similar to the subset of features considered in \cite{diffprivlib}. We sample 2000 training data points to construct graph $G$. The HELOC dataset is randomly split in a stratified fashion into 7403 training samples and 2468 test samples. The task is to predict risk associated with the home credit application. We selected 'ExternalRiskEstimate', 'MSinceOldestTradeOpen',
'AverageMInFile', 'NetFractionRevolvingBurden', and
'NetFractionInstallBurden' as predictor variables using recursive feature elimination. 25\% of positive and 25\% of negative samples are further sampled from the training and test datasets separately and the graph $G$ is constructed on 1851 sampled training dataset.

We train a DP logistic regression model for each of the datasets\cite{chaudhuri2011differentially}\footnote{\url{https://github.com/IBM/differential-privacy-library}} on the complete training datasets. Continuous attributes are scaled using a min-max scaler, assuming a prior knowledge of the minimum and maximum values for each feature set to bound the sensitivity of each feature \cite{near2021programming}. For any categorical features, we employ one-hot encoding. We get a test accuracy score of 80.5\% on the Adult dataset ($\epsilon_f=1, \delta_f=0$) and 70.54\% on the HELOC dataset ($\epsilon_f=3, \delta_f=0$). We note that training an accurate and private model is \textit{not} the focus of our work - our proposed solution assumes the availability of a black-box model with desirable performance.

\noindent
\subsection{Metrics.} We evaluate the efficacy of the differentially private recourse path by considering various distance measures and the quality of recourse. We consider a recourse path feasible if it passes through the denser region of the data manifold and lies closer to the data manifold. We propose to evaluate the realism of the recourse path\footnote{Similar to FACE, we consider recourse paths that pass through dense regions of the data as ideal, and assume that such a path would be realistic} using the metrics described below. 

\begin{itemize}
    \item 
    \texttt{PDensity}: This metric gives the measure of the total log-likelihood of the sample points constituting the recourse path under the density estimate of $D$. Let $\rho^D$ be the density estimate of $D$ and $\rho^{D}(x)$ denote the log-likelihood of the sample $x$ under density estimate $\rho^D$. For a recourse path \texttt{P} containing a sequence of points $\vectorbold{Z}_1, \vectorbold{Z}_2, \ldots, \vectorbold{Z}_p$ with $\vectorbold{Z}_p = \vectorbold{Z}^*$, \texttt{PDensity} is given by:
     \begin{equation}
       {
            \texttt{PDensity} = \frac{\sum_{i=1}^{p} \rho^D(\vectorbold{Z}_i)}{p}
            }
        \end{equation}

    \item
\texttt{PDistanceManifold}: This metric gives the distance of each step in $\texttt{P}$ from the training dataset. A Nearest
Neighbor (NN) algorithm is trained on the training dataset and
is used to get the \textit{l1-}distance between $\vectorbold{Z}_i$ and its nearest neighbour in training dataset\cite{verma2022amortized}.
    \begin{equation}\label{eqn:dist}
        {\texttt{PDistanceManifold} = \frac{\sum_{i=1}^{p}d_1(\vectorbold{Z}_i, NN(D,\vectorbold{Z}_i))}{p}
        }
    \end{equation} 
\end{itemize}


We consider average distance metrics for the recourse path. For any given distance function $d(.,.)$, we compute average distance of the recourse path {\tt P} as \texttt{PDistance} given by:
    \begin{equation}\label{eqn:dist}
        {\texttt{PDistance} = \frac{\sum_{i=1}^{p-1}d(\vectorbold{Z}_i, \vectorbold{Z}_{i+1}) + d(\vectorbold{Z},\vectorbold{Z}_1)}{p}
        }
    \end{equation}

We use L0 as  $d(.,.)$ in Eqn. \ref{eqn:dist} to report the sparsity of the recourse path and refer to it as PL0. 
We use L1 and L2 as  $d(.,.)$ in Eqn. \ref{eqn:dist} to report the proximity of the recourse path and refer to them as PL1 and PL2 respectively.

We additionally report the quality of CFE ($\vectorbold{Z}^*$) in terms of its sparsity (with respect to the original point) and its robustness for flipped prediction. We measure the robustness of the CFE by measuring the closeness of CFE to the favorable outcome data manifold (yNN) \cite{pawelczyk2021carla} by counting the instances in its neighborhood with the same outcome. We measure the sparsity of the CFE by computing the number of feature changes not necessary for change in prediction (Redundancy)\cite{pawelczyk2021carla}. While the primary metric of focus for us is based on feasibility, each of the above measures can help inform on the actionability (the actionability of the recourse path is inherent to the proposed solution as the graph can be constructed with constraints) and feasibility of the recourse path in different ways. We also report the average time it takes to find the recourse path.

\subsection{Impact of privacy.}
We evaluate PrivRecourse against two approaches - baseline approach that simply perturbs the data, and the approach using DP synthetic data to form the graph - in Tab. \ref{tab:results}. For the experiments in Tab. \ref{tab:results}, we consider a fully connected graph on the sampled train set. The query instances are the data points with unfavorable outcomes from the sampled test set for the experiments.

\begin{table*}[h]
    \centering
    \caption{Recourse path for query 1 from Adult dataset with non-private FACE algorithm, DP-SD and PrivRecourse. The columns are  The first row, for each method, represents the query instance, the second row represents the nearest point in the published graph with the same class label (hence, it takes values close to the  point in the original data for FACE and DP data for others, and that can be greater than or less than the query feature value) and the last row represents the CFE.}
   \scalebox{1}{
    {\begin{tabular}{c | c c c c c}
           & age & education-num & capital-gain & capital-loss & hours-per-week\\
         \toprule
        \multirow{3}{*}{FACE} & 41 & 10  & 0 & 0 & 40\\
        & 43  & 10 & 0 & 0 & 40\\
        & 46 & 10 & 0 & 0 & 40\\
        & 48 & 10 & 0 & 0 & 24\\
         \midrule
         \multirow{3}{*}{PrivRecourse} & 41 & 10  & 0 & 0 & 40\\
        & 46 & 10 & 0 & 0 & 40\\
        & 52 & 10 & 0 & 0 & 24\\
    \end{tabular}
    }       
    \label{tab:recoruse_path}
        }
\end{table*}

\noindent
\paragraph{Baseline.} The simple and naive approach is to perturb the individual samples such that they satisfy differential privacy. We apply Laplace noise with sensitivity of 1 to continuous attributes after clipping the values based on domain knowledge. The privacy budget is equally split per feature per sample. The noisy data with ($\epsilon_k,\delta_k$)-DP guarantees is published to further construct the graph (line 3--18 in Alg. \ref{alg:main})


\noindent
\paragraph{DP Synthetic Data (DP-SD).} We use the MST algorithm\cite{tao2021benchmarking}\footnote{\url{https://github.com/ryan112358/private-pgm}} with ($\epsilon_k,\delta_k$) as the privacy parameters and other default parameters to generate synthetic data that has the same number of rows as the sampld training data. The average error for generating synthetic data for the selected feature attributes for Adult dataset was 0.19 and for HELOC dataset was 0.21. The published ($\epsilon_k,\delta_k$)-DP synthetic data is published to further construct the graph (line 3--18 in Alg. \ref{alg:main})


\noindent
\paragraph{PrivRecourse.}
Clustering is performed with 11 iterations and internalK of 5. The privacy budget is equally split over total iterations. i.e.~ for every iteration $\epsilon_p = \epsilon_k/10$ , and for every iteration $0.75\epsilon_p$ is used for the exponential mechanism and $0.25\epsilon_p$ for the Laplace mechanism. The cluster centers are published with ($\epsilon_k,\delta_k$)-DP to further construct the graph (line 3--18 in Alg. \ref{alg:main})

All the methods in Tab. \ref{tab:results} utilize the same model for each dataset, ensuring a fair comparison across methods with a shared decision boundary. We compare the feasibility of the recourse path with respect to the original training data across all methods. We compute $\rho^D$ using the Gaussian kernel density estimate. The average runtime reports the average time\footnote{All experiments were run on 2.6 GHz 6-Core Intel Core i7} taken to generate the recourse path for each query. We note the ML model can be trained on the published noisy or synthetic data to allot more privacy budget to the data publishing algorithms. We think this can hurt the utility of the ML model if the DP-published data does not capture the distribution of the original data well. To have the same privacy budget across all methods, overall $\epsilon_k=1,\delta_k=0$ is used. We compare the feasibility of the recourse path with respect to the original training data across all methods. We compute $\rho^D$ using the Gaussian kernel density estimate. The average runtime reports the average time\footnote{All experiments were run on 2.6 GHz 6-Core Intel Core i7} taken to generate the recourse path for each query.

PrivRecourse performs better in terms of feasibility ({\tt PDensity} and {\tt PDistanceManifold}) metrics across all datasets as the generated cluster centers can represent the denser regions of the dataset while being closer to the original data manifold.  DP-SD gives relatively lower feasible paths, with the baseline performing the worst across all datasets. DP-SD performs better in terms of proximity and sparsity across all datasets. This is consistent with the inherent trade-off between being realistic with respect to the data manifold and finding points which are close to the original point (explored previously in the CFE literature \cite{van2021interpretable}). We do note that the metrics and the CFE generated depend on the dataset, its distribution and the attributes chosen. Baseline performs worst for all the metrics demonstrating that naive solution can negatively impact the data distribution.

\subsubsection{Recourse Example with constraints.} 
We provide a recourse example for the Adult dataset when we consider constraints that the values of age and education-num should not decrease when constructing the graph with a \textit{l2}-distance threshold of 0.4. We observe that the constraints might not lead to a fully connected graph and can affect the success rate of finding CFE. The success rate drops down to 96\% for FACE and 94\% with the clustering approach. We show the recourse path generated for one of the examples in Tab. \ref{tab:recoruse_path} with FACE and with PrivRecourse.
The first row, for each method, represents the query instance, the second row represents the nearest point in the published graph with the same class label (hence, it takes values close to the  point in the original data for FACE and DP data for others, and that can be greater than or less than the query feature value) and the last row represents the CFE. We observe that constraints hold across all methods. PrivRecourse generates recourse points that represent the data manifold well and ensures that perturbations in the example will not reverse the outcome. We do note that because the constraints are not set on all features, the generated recourse can require human intervention for correctness and actionability. Moreover, the generated recourse is highly influenced by the training dataset and the distance threshold considering the endogenous nature of the FACE algorithm, and the quality of the recourse can vary across examples. Nevertheless, these recourse paths can be used to assist human intervention. In the example in Tab. \ref{tab:recoruse_path} the path generated by PrivRecourse has a higher age. We think this is due to the clustering approach. We however note that the recommended multi-step recourse recommendation remains same - "increase in age and then decrease in hours-per-week".

\subsection{Discussion and Limitations}
 Our proposed pipeline considers the shortest path returned by the algorithm as the recourse path. We note that multiple recourse paths can be returned to satisfy diversity by returning the second, third, and fourth shortest paths as multiple recourses. Additional constraints can be employed while generating these paths to see changes in different features. The advantage of a privacy-published graph is that a synthetic node or a noisy new observed data point can be added at any time to improve the quality of generated recourse paths and additional constraints can be imposed to modify the graph. With additional constraints, the graph may lead to more than one connected component impacting the success rate of finding the recourse path. Human intervention may be required while adding proper constraints and also validating the actionability of the recourse paths, just as in any algorithmic recourse method. 

 Of the considered approaches in Tab. \ref{tab:results}, one might choose either DP-SD or PrivRecourse for feasibility based on the required properties for the CFE and recourse path to be generated, and the ability to generate synthetic data. We note that the clusters will represent the data manifold well, depending on the clustering algorithm chosen to fit in the proposed pipeline, and we find that this proposed approach works well. Additionally, synthetic data may be hard to generate for every dataset. However, we additionally note that any number of data points can be sampled using the DP synthetic data generator, which can yield different results. If time is the constraint, less number of data points can be generated or if more dense representation is desired, more number of points can be generated. For a fair comparison, we consider the same number of data points as the original data in our empirical analysis. We introduce PrivRecourse as an initial method that can be used for providing privacy-preserving realistic recourse that can motivate further research in this domain.

\section{Conclusion}
We propose a privacy-preserving pipeline, PrivRecourse, to generate recourse paths that are realistic - feasible, and actionable. PrivRecourse, to the best of our knowledge, is the first to explore privacy-preserving recourse paths. PrivRecourse leverages privacy-preserving clustering to represent the private version of the dataset, and then forms a graph on these cluster centers to get recourse paths. We empirically evaluate PrivRecourse against two baseline approaches and find that PrivRecourse generates recourse paths that lie close to the observed data manifold. We consider PrivRecourse as a starting point for providing private recourse paths, opening up interesting research directions in this problem domain. For future work, we will build on PrivRecourse to provide recourse paths that are fair with respect to sensitive attributes, more robust, and with better utility.

\section*{Disclaimer}
This paper was prepared for informational purposes 
by the Artificial Intelligence Research group of JPMorgan Chase \& Co and its affiliates (“J.P. Morgan”) and is not a product of the Research Department of J.P. Morgan.  J.P. Morgan makes no representation and warranty whatsoever and disclaims all liability, for the completeness, accuracy or reliability of the information contained herein.  This document is not intended as investment research or investment advice, or a recommendation, offer or solicitation for the purchase or sale of any security, financial instrument, financial product or service, or to be used in any way for evaluating the merits of participating in any transaction, and shall not constitute a solicitation under any jurisdiction or to any person, if such solicitation under such jurisdiction or to such person would be unlawful.  © 2023 JPMorgan Chase \& Co. All rights reserved 

\bibliographystyle{ACM-Reference-Format}
\bibliography{ref}
\end{document}